%% file: root.tex
\documentclass[letterpaper, 10 pt, journal, twoside]{IEEEtran} 
\input{preamble}
\IEEEoverridecommandlockouts                              

\title{\LARGE \bf
\acronym: Scaling Up Natural Language Understanding\\ for Multi-Robots Through Hierarchical Temporal Logic\\ Task Representation
}

\author{Shaojun~Xu$^{1}$$^{\,,}$${^*}$$^{\,,}$$^{\dagger}$, Xusheng~Luo$^{2}$$^{\,,}$${^*}$, Yutong~Huang$^{2}$, Letian~Leng$^{2}$, Ruixuan Liu$^{2}$, Changliu Liu$^{2}$
\thanks{${^*}$Equal contribution.}
\thanks{$^{1}$Shaojun Xu is with Department of Precision Instrument, Tsinghua University, Beijing, 100084, China {\tt\small xusj24@mails.tsinghua.edu.cn}}%
\thanks{$^{2}$Xusheng Luo, Yutong~Huang, Letian~Leng, Ruixuan Liu and Changliu Liu are with Robotics Institute, Carnegie Mellon University, Pittsburgh, PA 15213, USA {\tt\small\{xushengl,  yutongh3,  lleng,  ruixuanl, cliu6\}@andrew.cmu.edu}}%
\thanks{$^{\dagger}$Shaojun Xu was an intern at CMU when this work was conducted.}
}

\begin{document}

\maketitle
\thispagestyle{empty}
\pagestyle{empty}

\begin{abstract} 

To enable non-experts to specify long-horizon, multi-robot collaborative tasks, language models are increasingly used to translate natural language commands into formal specifications. However, because translation can occur in multiple ways, such translations may lack accuracy or lead to inefficient multi-robot planning. Our key insight is that concise hierarchical specifications can simplify planning while remaining straightforward to derive from human instructions. We propose~\acronym{}, a framework that translates natural language commands into hierarchical Linear Temporal Logic (LTL) and solves the corresponding planning problem. The translation involves two steps leveraging Large Language Models (LLMs). First, an LLM transforms instructions into a Hierarchical Task Tree, capturing logical and temporal relations. Next, a fine-tuned LLM converts sub-tasks into flat LTL formulas, which are aggregated into hierarchical specifications, with the lowest level corresponding to ordered robot actions. These specifications are then used with off-the-shelf planners. Our~\acronym{} demonstrates the potential of LLMs in hierarchical reasoning for multi-robot task planning. Evaluations in simulation and real-world experiments with human participants show that~\acronym{} outperforms existing methods, handling more complex instructions while achieving higher success rates and lower costs in task allocation and planning. Additional details are available at \href{https://nl2hltl2plan.github.io/}{nl2hltl2plan.github.io}.
\end{abstract}


\input{introduction}
\input{literature}

\input{hltl}

\input{nl2tl}
\input{simulation}

\input{conclusion}
\begin{sizeddisplay}{\footnotesize}
\bibliographystyle{IEEEtran}
\bibliography{xl_bib_2}
\end{sizeddisplay}
\end{document}

%% file: preamble.tex
\usepackage{hyperref}
\usepackage[T1]{fontenc}
\usepackage[utf8]{inputenc}
\usepackage{tabularx,ragged2e,booktabs,caption}
\newcolumntype{C}[1]{>{\Centering}m{#1}}

\usepackage{graphicx} 
\graphicspath{{figs/}}
\usepackage{epsfig} 
\usepackage{ragged2e}
\usepackage{amsmath}
\usepackage{amssymb}
\usepackage{epstopdf}
\usepackage{cite}
\usepackage[noend,ruled,linesnumbered]{algorithm2e}
\SetKwComment{Comment}{$\triangleright$\ } {}
\usepackage{multirow}
\usepackage{rotating}
\usepackage{subfigure}
\usepackage{color}
\usepackage{mysymbol}
\usepackage{url}
\usepackage{ltlfonts}	
\usepackage{ulem}
\usepackage{enumerate}
\usepackage{stmaryrd}
\usepackage{array}
\usepackage[most]{tcolorbox}
\usepackage{listings}
\usepackage{float}
\usepackage{cuted}
\usepackage[pdf]{graphviz}
\usepackage{svg} 
\usepackage{wrapfig}    
\usepackage[numbers,sort&compress]{natbib}
\usepackage[font=footnotesize]{caption}
\allowdisplaybreaks

\newtheorem{theorem}{Theorem}[section]

\newtheorem{exmp}{Example}

\newtheorem{defn}[theorem]{Definition}
\newtheorem{rem}[theorem]{Remark}

\newtheorem{definition}[theorem]{Definition}

\newcommand{\level}[2]{\phi_{#2}^{#1}}

\usepackage{enumitem}
\setlist[enumerate]{leftmargin=*}

\renewcommand{\theenumi}{\arabic{enumi}}
\renewcommand{\theenumii}{\theenumi.\arabic{enumii}}
\renewcommand{\theenumiii}{\theenumii.\arabic{enumiii}}
\renewcommand{\theenumiv}{\theenumiii.\arabic{enumiv}}

\newcommand{\dish}[2]{\pi_{\text{#1}}^{#2}}

\lstdefinelanguage{json}{
    basicstyle=\normalfont\ttfamily,
    numbers=left,
    numberstyle=\scriptsize,
    stepnumber=1,
    numbersep=8pt,
    showstringspaces=false,
    breaklines=true,
    frame=single,
    backgroundcolor=\color{white},
    stringstyle=\color{blue},
    keywordstyle=\color{red},
    commentstyle=\color{gray},
    morestring=[b]",
    morestring=[d]',
    morecomment=[l]{//},
    morecomment=[s]{/*}{*/},
    morecomment=[l]{\#},
    morekeywords={null,false,true}
}

\definecolor{CMURed}		        {RGB}{145,11,46}	      

\newcommand{\bhline}{\noalign{\hrule height 1.2pt}}

\newenvironment{sizeddisplay}[1]
 {\par\nopagebreak#1\noindent\ignorespaces}
 {\nopagebreak\ignorespacesafterend}

 \usepackage{titlesec}

\titleformat{\paragraph}[runin] 
  {\normalfont\normalsize\it} 
  {\theparagraph} 
  {0.5em} 
  {} 
  [:] 
\titlespacing{\paragraph}
  {1em} 
  {0em} 
  {0.5em} 

\renewcommand{\theparagraph}{\arabic{paragraph})}

\newcommand{\acronym}{\textsc{Nl2Hltl2Plan}}
\newcommand{\acronymsmartllm}{\textsc{Smart-Llm}}

%% file: introduction.tex
\section{Introduction}
Large Language Models (LLMs), trained on vast text corpora, display common sense reasoning abilities that enable them to handle routine tasks expressed in human language. The development of LLMs has opened up accessible ways for non-experts to instruct and interact with robots through natural language~\cite{padalkar2023open}. One approach is the~\textbf{neuro-symbolic paradigm}~\cite{belle2023neuro}, in which an intermediate formal specifiaction is derived from natural language input and subsequently used by existing solvers for planning, offering a structured and consistent interpretation of tasks~\cite{cohen2024survey}. This approach is also data-efficient, especially considering the limited availability of robotic data. A core requirement in this line of work is that specifications should be accurate, concise and enhance the downstream planners’ effectiveness. It is widely recognized that hierarchical models outperform flat models in interpretability and efficiency~\cite{tenenbaum2011grow, kemp2007learning}. However, effectively incorporating human-derived hierarchical insights into algorithms necessitates careful engineering, posing a challenge to leveraging hierarchical planners.


\begin{figure}[!t]
    \centering
    \includegraphics[width=\linewidth, trim=0.2cm 0cm 0cm 0cm, clip]{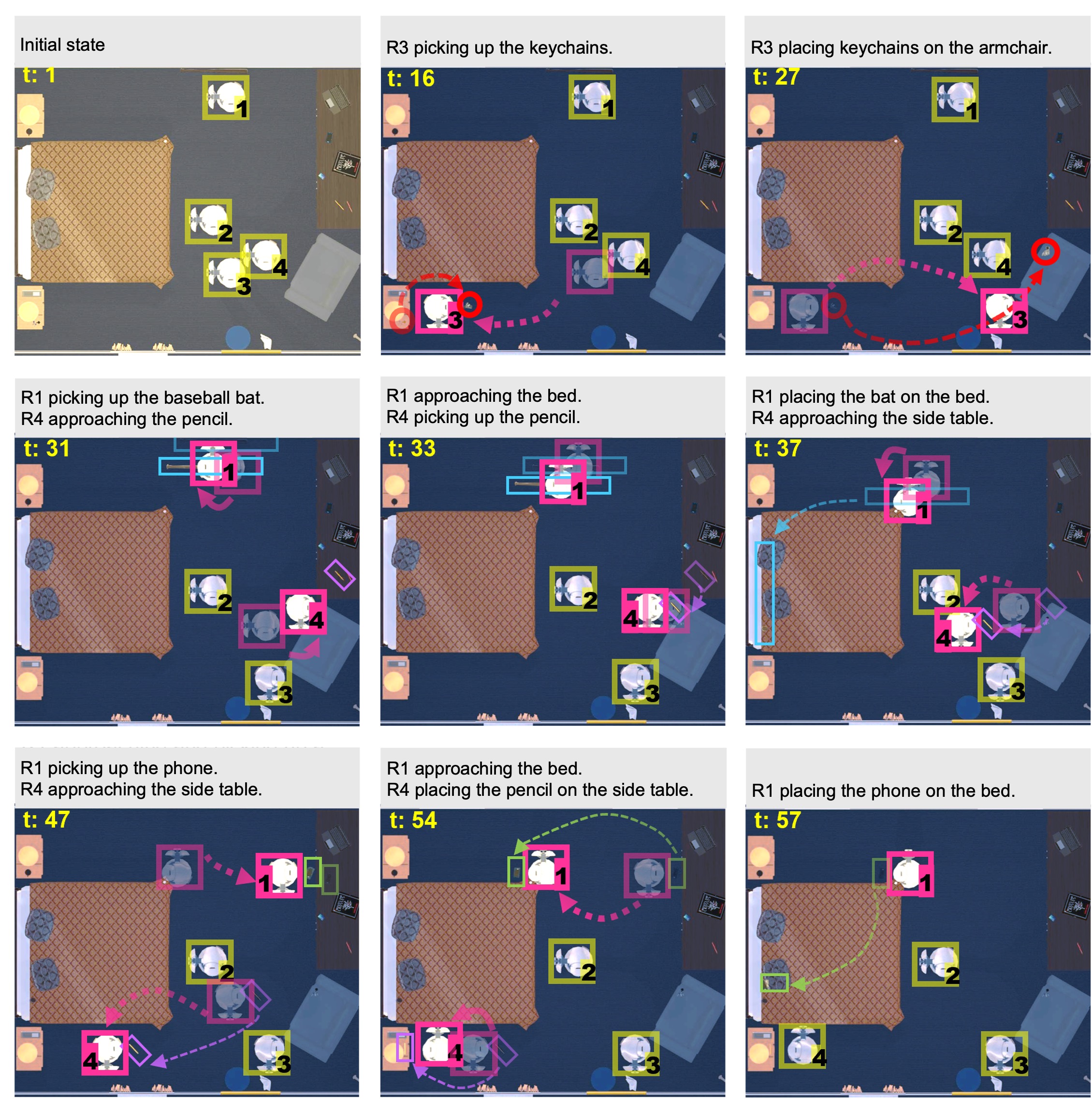}
    \caption{A sequence of images, arranged from left to right and top to bottom, depicts the task 
    ``First, put a set of keychains on the armchair. Retrieve a pencil and put it on the side table. Move the phone and the bat to the bed in any order'', objects and their trajectories are marked with different colors as follows, keychains (red), bat (blue), pencil (purple) and phone (green). $t$ represents the discrete time steps in simulation.
    }
    \label{fig:ai2thor} 
    \vspace{-10pt}
\end{figure}

\begin{figure*}
    \centering
    \includegraphics[width=\linewidth]{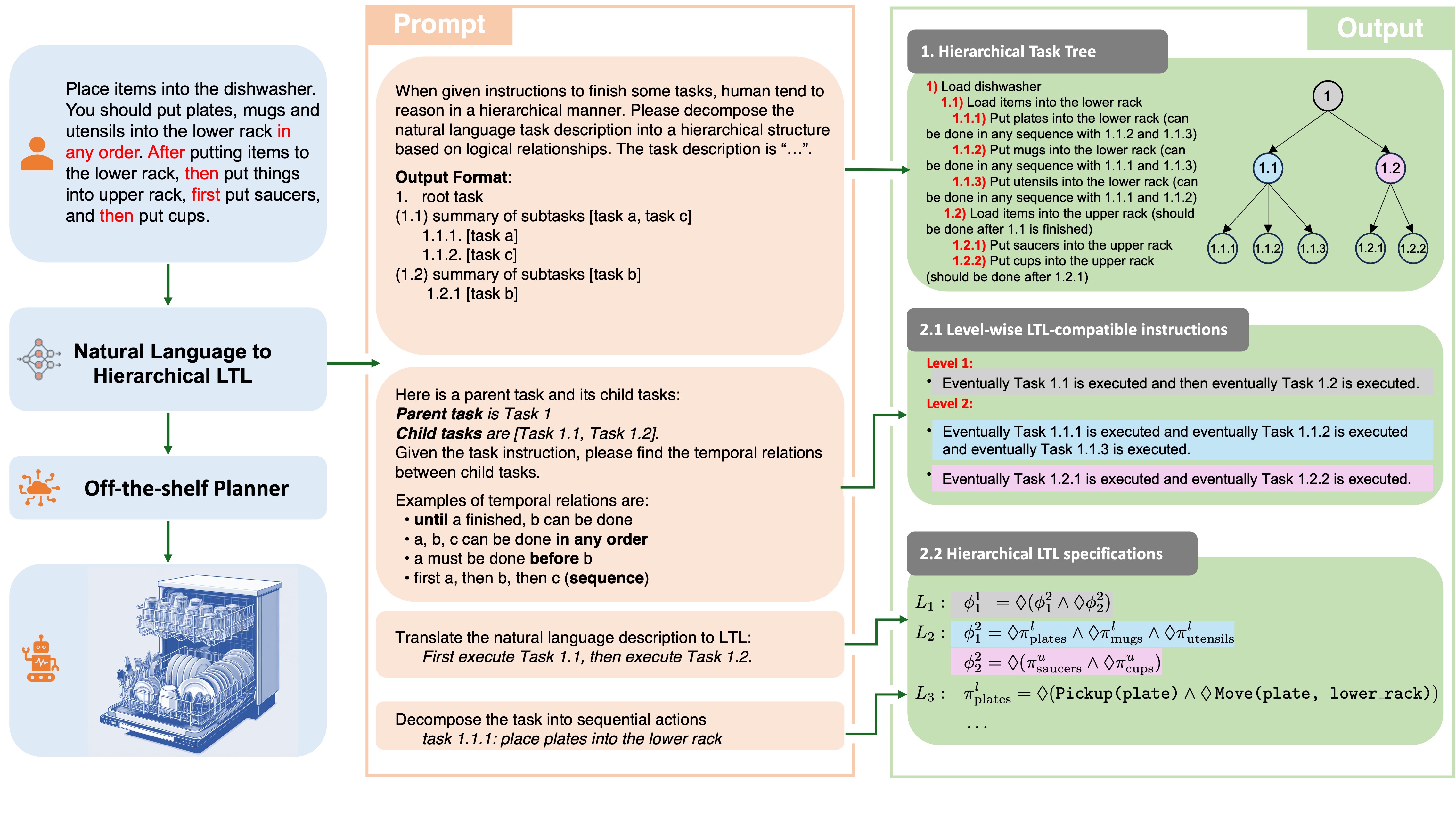}
    \caption{Overview of the framework~\acronym. The non-leaf nodes in the Hierarchical Task Tree (see Section~\ref{sec:htt}), the language descriptions of subtasks, and the flat specifications are color-coded to indicate one-to-one correspondence. Summary snippets of the prompts are provided, with more information accessible on the project page~\href{https://nl2hltl2plan.github.io}{nl2hltl2plan.github.io}.}
    \label{fig:overview}
    \vspace{-10pt}
\end{figure*}

Recently,~\cite{liembodied} introduced the use of Linear Temporal Logic (LTL) as a standardized framework for specifying goals in embodied decision making, highlighting its expressiveness and compactness. Nonetheless, their approach is limited to addressing simple goals defined by an ordered sequence. Our insight is that task hierarchy can be progressively obtained from language instruction with the help of a LLM. 
In light of the above, we propose harnessing LLMs as language-based task hierarchy extractors. 
Hierarchical Linear Temporal Logic, a variant of formal languages introduced in work~\cite{luo2024simultaneous}, is adopted as intermediate task specification, which is succinct and results in efficient planners compared to its flat counterpart, aligning well with hierarchically represented human instructions and has been applied to robotics~\cite{luo2021abstraction,luo2022temporal}.  With hierarchy extraction, we can handle multiple-sentences  instructions involving multiple robots, while related work primarily focuses on short instructions for single robot.

Via fine-tuned LLM, a naive approach to converting language instructions directly into hierarchical LTL is easy to implement. However, this technique tends to perform poorly as LLMs are still not good at logical reasoning~\cite{xu2023large}, which is crucial for crafting logical formulas. Furthermore, LTL formulas in the dataset for learning translations generally have between 2 and 4 propositions~\cite{liu2023grounding}, rendering them unsuitable for instructions that involve multiple lengthy sentences.
We propose a two-step approach~\acronym{} to unlock the expressive prowess of temporal logic, converting instructions into hierarchical LTL. Initially, upon receiving an instruction, we prompt an LLM to generate and gradually refine a task representation which is a simplified version of Hierarchical Task Network~\cite{ghallab2016automated}. 
Subsequently, in the second phase, sub-tasks of each task can be translated into a single flat LTL via a fine-tuned LLM. Through iterative processing of all sub-tasks of every task in the intermediary phase, we can construct hierarchical LTL specifications,  where the lowest level corresponds to sequentially ordered robot actions.
This paradigm of using a formal representationis is data efficient and interpretable~\cite{cohen2024survey}.

With~\acronym, human instructions are ready for use by off-the-shelf  hierarchical LTL planners, and applied to multi-robot systems with specified objective like cost optimization, which differs from most works that only consider finding feasible solutions rather than optimize under specific objectives. The translation of hierarchical task instructions into hierarchical LTL proves to be more straightforward and dependable compared to translating into a cumbersome flat formula, a challenge not solved by existing works~\cite{liu2023grounding,chen2023nl2tl,cosler2023nl2spec}.


{\bf Contributions:} 1) We proposed a neuro-symbolic method~\acronym{} to extract task hierarchies from instructions to facilitate multi-robot planning for long-horizon tasks; 2) We developed a method that transforms language into hierarchical LTL, thus integrating human-derived hierarchical knowledge in planning solvers; 3) We validated our method through simulations and real-world experiments using instructions to formulate plans for multi-robot mobile manipulation tasks.


%% file: literature.tex
\section{Related Work}

{\bf Language-Conditioned Robotic Planning:}
Given instructions, there are two primary methods for generating actions~\cite{cohen2024survey}. The first uses deep-learning techniques to translate instructions into low-level actions, such as joint states. Systems 
 on this have shown capabilities across multiple modalities ~\cite{padalkar2023open, team2024octo, jiang2022vima,li2023vision}, but they depend on large volumes of data. 
Others translate instructions into an intermediate representation, then employing off-the-shelf solvers to generate actions, which limits the solution space, further reducing the need for extensive data. 
The intermediate representations employed can vary from formal planning formalisms such as Planning Domain Definition Language (PDDL) and temporal logics, to less formal structures like code or predefined skills.

LLMs have been used to extract goal states and domain descriptions from instructions via prompting~\cite{xie2023translating,liu2023llm+,valmeekam2024planbench}.
Their capacity to generate low-level code or call APIs has been verified~\cite{singh2023progprompt,liang2023code,huang2023voxposer,kannan2023smart,hu2024deploying}. An updatable skill library, instead of calling fixed APIs, are introduced by Voyager~\cite{wang2023voyager} and \textsc{Saycan}~\cite{brohan2023can}, and enhanced by \textsc{InnerMonologue}~\cite{huang2023inner}, \textsc{KnowNo}~\cite{ren2023robots} through integrating feedback or help seeking ability.
A commonality is their focus on single-robot scenarios, however, extension to multi-robot scenarios remains largely unexplored.

{\bf Natural Language to Temporal Logic:}
Early attempts at translating natural language into temporal logics relied on grammar-based methods, which excel at processing structured inputs~\cite{konrad2005real}. Recently, the use of LLMs for this task has gained traction, leveraging tools like GPT to generate LTL formulas~\cite{fuggitti2023nl2ltl, cosler2023nl2spec}. While these approaches focus on the translation process, they often overlook the critical issue of grounding language in robotics—linking linguistic instructions to physical actions and environments. To address this,~\cite{pan2023data} fine-tuned an LLM using a synthetic dataset that pairs natural language instructions with temporal logic formulas designed for quadrotor tasks. Similarly, weakly supervised semantic parsers have been developed to learn from execution trajectories without requiring explicit LTL annotations~\cite{patel2019learning,wang2021learning}. Systems such as \textsc{Lang2LTL}~\cite{liu2023lang2ltl}, \textsc{NL2TL}~\cite{chen-etal-2023-nl2tl}, and others~\cite{hsu2024s} employ LLMs to convert domain-specific commands (e.g., for navigation or motion planning) into formal specifications. In contrast,~\cite{wang2023conformal} adopts a predefined LTL specification approach, where predicates are defined using succinct human instructions. Our \acronym{} extends these 
capabilities, supporting more complex specifications with over 10 atomic propositions and enabling task allocation across multiple robots—surpassing the scope of prior works, which typically handle fewer than five atomic propositions.

{\bf LLMs to Multi-Robots:}
To tackle the problem, a notable trend in adapting LLMs for use in multi-robot systems is raising.~\acronymsmartllm~\cite{kannan2023smart} uses an LLM to synthesize code that facilitates task decomposition, coalition formation, and task allocation. 
Multiple intermediate approaches have been implemented in multi robot planning, such as dialogue-based framework~\cite{mandi2023roco}, behavior trees~\cite{lykov2023llm}, batch of multi communication frameworks (centralized, decentralized, or hybrid) \cite{chen2023scalable}, and address deadlock resolution in navigation scenarios~\cite{garg2024large}. Decentralized LLM-based planner~\cite{wang2024safe} and global LLM-based planners~\cite{yu2023co} are introduced to enhance the efficiency of target searches or make individual decisions autonomously
However, the works mentioned above focus on finding feasible solutions. 
In contrast, our research can optimize the cost and time required to complete tasks.

%% file: hltl.tex
\section{Hierarchical Linear Temporal Logic}\label{sec:preliminaries}
Linear Temporal Logic (LTL) is composed of basic statements, referred to as atomic propositions $\mathcal{AP}$, along with boolean operators such as conjunction ($\wedge$) and negation ($\neg$), temporal operators like next ($\bigcirc$) and until ($\mathcal{U}$)~\cite{baier2008principles}:
\begin{align}\label{eq:grammar}
    \phi:=\top~|~\pi~|~\phi_1\wedge\phi_2~|~\neg\phi~|~\bigcirc\phi~|~\phi_1~\mathcal{U}~\phi_2, 
\end{align}
where $\top$ stands for a true statement, and $\pi$ is a boolean valued atomic proposition. Other temporal operators can be derived from $\mathcal{U}$, such as $\Diamond \phi$ that implies $\phi$ will be true at a future time.
We focus on a subset of LTL known as syntactically co-safe formulas (sc-LTL)~\cite{kupferman2001model}. Any LTL formula encompassing only the temporal operators $\Diamond$ and $\mathcal{U}$ and written in positive normal form (where negation is exclusively before atomic propositions) is classified under sc-LTL formulas~\cite{kupferman2001model}, which can be satisfied by finite sequences followed by any infinite repetitions. This makes sc-LTL apt for reasoning about robot tasks with finite duration.

\begin{defn}[Hierarchical sc-LTL~\cite{luo2024simultaneous}]\label{def:hltl}
Hierarchical sc-LTL is structured into $K$ levels, labeled as $L_1, \ldots, L_K$, arranged from the highest to the lowest.   Each level $L_k$, where $k \in [K]$, contains $n_k$ sc-LTL formulas.  The hierarchical sc-LTL can be represented as $\Phi = \left\{\level{i}{k} \,|\, k \in [K], i \in [n_k]\right\}$, where $\level{i}{k}$ denotes the $i$-th sc-LTL formula at level $L_k$. The hierarchical sc-LTL adheres to the following rules:
\begin{enumerate}
    \item Each formula at a given level $L_k$, for $k \in [K-1]$, is derived from the formulas at the next lower level $L_{k+1}$.
    \item \label{cond:once-higher} Every formula at any level other than the highest (i.e., $k = 2, \ldots, K$) is included in exactly one formula at the next higher level $L_{k-1}$.
    \item \label{item:hltl_3} Atomic propositions are used exclusively within the formulas at the lowest level $L_K$.
\end{enumerate}
\end{defn}

Let $\Phi^k$ denote the set of formulas at level $L_k$ with $k \in [K]$. We refer to each specification $\level{k}{i}$ in $\Phi$ as the ``flat'' specification, which can be organized in a tree-like specification hierarchy graph, where each node represents a flat sc-LTL specification. Edges between nodes indicate that one specification belongs to another as a {\it composite proposition}. The $K$-th level leaf nodes represent {\it leaf specifications} that consist only of atomic propositions, while non-leaf nodes represent {\it non-leaf specifications} made up of composite propositions.

\begin{exmp}[Dishwasher Loading Problem]\label{exmp:dishwasher}
Consider the following instruction: ``Place items into the dishwasher. Put plates, mugs and utensils into the lower rack {\it in any order}. {\it After} putting items to the lower rack, {\it then} put things into upper rack, {\it first} put saucers, {\it and} then put cups.'' The hierarchical LTL is:
\begin{align}\label{eq:hier}
     L_1: \quad &  \level{1}{1}\;\, = \Diamond (\level{1}{2} \wedge \Diamond \level{2}{2})  \nonumber \\
     L_2: \quad &  \level{1}{2} = \Diamond \dish{plates}{l}  \wedge \Diamond  \dish{mugs}{l} \wedge \Diamond  \dish{utensils}{l} \\
                & \level{2}{2} = \Diamond (\dish{saucers}{u} \wedge \Diamond \dish{cups}{u}), \nonumber
\end{align}
where $\level{1}{2}$ and $\level{2}{2}$ are {composite propositions}, and the formula $\Diamond (\level{1}{2} \wedge \Diamond \level{2}{2})$ specifies that $\level{1}{2}$ should be fulfilled before moving on to $\level{2}{2}$. $\dish{i}{j}$ represents atomic propositions, denoting the act of placing a specific type of dishware. Note that the lowest level $L_2$ only includes atomic propositions.
\end{exmp}


%% file: nl2tl.tex
\section{Methodology:~\acronym}
LLMs excel in common sense reasoning but perform poorly in logical reasoning and lack grounding in the available robot skills~\cite{petroni2019language,xu2023large}. Therefore, we propose a two-stage method for translating natural language into hierarchical LTL using an intermediary structure known as the Hierarchical Task Tree. 
\subsection{Conversion from instructions to Hierarchical Task Tree}\label{sec:htt}
\begin{definition}[Hierarchical Task Tree (HTT)]\label{def:htt}
A Hierarchical Task Tree (HTT) is a tree $\ccalT = (\ccalV, \ccalE, \ccalR)$, where 
\begin{itemize}
    \item $\ccalV=\{v_1,v_2,\ldots,v_n\}$ denotes the set of nodes. Each node is associated with an instruction of its respective task;
    \item $\ccalE \subseteq \ccalV \times \ccalV$ represents the edges, indicating a decomposition relationship between tasks. An edge $e = (v_1, v_2) \in \ccalE$ implies that {\it child} task $v_2$ is in sub-tasks set of {\it parent} task $v_1$. The node set $\ccalV$ can be partitioned into multiple disjoint subsets $\{\ccalV_1, \ldots, \ccalV_m\}$, such that all nodes within the same subset $\ccalV_i$ share the same parent node. 
    \item  $\ccalR \subseteq \ccalV \times \ccalV$ defines the set of temporal relations between {\it sibling} tasks, which are decompositions of the same parent task. A relation $(v_1, v_2) \in \ccalR$, where $v_1, v_2 \in \ccalV_i$ for some $i \in \{1, \ldots, m\}$, indicates that task $v_1$ should be completed before task $v_2$.
\end{itemize}

\end{definition}

The HTT is a simplified version of the hierarchical task network (HTN) as it is specifically designed to align with the structure of hierarchical LTL. The tree unfolds level by level, where each child task is a decomposition of its parent task. The relation $\ccalR$ specifically captures the temporal relationships between sibling tasks that share the same parent. The temporal relationship between any two tasks can be inferred by tracing their lineage back to their common ancestor. The primary distinction between HTT and HTN is that HTN includes interdependencies between sub-tasks under different parent tasks and each node in the HTT is solely focused on the sub-task goal and does not incorporate other properties like preconditions and effects that are found in HTN. A LLM is employed to construct the HTT through a two-step process from given task instruction, as outlined in step 1 of Fig.~\ref{fig:overview}. 

\paragraph{HTT without temporal relations $\ccalR$ }
The first step involves generating the nodes $\ccalV$ and edges $\ccalE$, excluding the temporal relations $\ccalR$. The LLM is employed to decompose the whole task into a structured hierarchy and the decomposition continues until a task consists solely of sequential operations performed on a single object. 

\paragraph{Add temporal relations $\ccalR$ }
For each non-leaf node $v$, we consider $\ccalV'$, which represents its child tasks at the level directly beneath it. Then temporal relations between sibling tasks within $\ccalV'$ is determined by LLM. 
\begin{algorithm}[!t]
\caption{Construction of hierarchical LTL}
\LinesNumbered
\label{alg:construction}
\KwIn {HTT $\ccalT$}
\KwOut {Hierarchical LTL specifications}
$\ccalV_{\text{front}} = \varnothing$, $\Phi = \varnothing$ \Comment*[r]{{\color{blue}$\ccalV_{\text{front}}$ is a stack that contains nodes to be expanded}}
$\ccalV_{\text{front}}.\texttt{push}(v_{\text{root}})$  \Comment*[r]{{\color{blue} Add root node}}
\While{ $\ccalV_{\text{front}} \neq \varnothing$ } {
    $v = \ccalV_{\text{front}}.\texttt{pop}()$\;
     $k = \texttt{GetDepth}(v)$ \Comment*[r]{{\color{blue}Get the depth of node $v$ in $\ccalT$, $\texttt{GetDepth}(v_{\text{root}})=1$}}
    $i = \texttt{Count}(\Phi, k)$ \Comment*[r]{{\color{blue}Count the number of specifications at level $k$ in $\Phi$}}
    \If{$v$ is a leaf node}{
        $\phi_{i+1}^k$ = \texttt{ActionCompletion}($v$)\;
    }
    \Else{
    $\ccalV' = \texttt{GetChildren}(\ccalT, v)$\;
    $\ccalV_{\text{front}}.\texttt{push}(\ccalV')$ 
    $\ccalR' = \texttt{GetTemporalRelations}(\ccalT, \ccalV')$\;
    $\phi_{i+1}^k = \texttt{GenerateLTL}(\ccalV', \ccalR')$ \Comment*[r]{{\color{blue}Generate the single LTL}}
    }
    $\Phi.\texttt{add}(\phi_{i+1}^k)$\;  
}
\Return $\Phi$\;
\end{algorithm}
\vspace{-10pt}
\subsection{Generation of task-wise flat LTL specifications}\label{sec:translation}
Once the HTT representation is obtained, a flat LTL is generated for each node via a breadth-first search; see Alg.~\ref{alg:construction}. 
\paragraph{Logical search} 
For every non-leaf node $v$, we gather its child tasks $\ccalV'$ and the temporal relations among them, defined by $\ccalR' \subseteq \ccalV' \times \ccalV'$. We then use an LLM to rephrase these child tasks with their temporal relations into syntactically correct sentences aligned with the semantics of LTL specifications (as illustrated in step 2.1 in Fig.~\ref{fig:overview}). A fine-tuned LLM is then used as a translator to obtain single LTL formula from reformulated sentences (as depicted in step 2.2 in Fig.~\ref{fig:overview}). To this end, we first developed a dataset comprising pairs of  natural language descriptions and their corresponding LTL formulas, and then fine-tune a language model for translation, \verb|Mistral-7B-Instruct-v0.2|~\cite{jiang2023mistral}. Training datasets were synthesized from sources including Efficient-Eng-2-LTL~\cite{pan2023data}, Lang2LTL~\cite{liu2023lang2ltl}, nl2spec~\cite{cosler2023nl2spec}, and NL2TL~\cite{chen-etal-2023-nl2tl}. Given the domain-specific nature of these datasets, we substituted specific tasks with generic symbols such as ``task 1.1 should be completed before task 1.2'' paired with the LTL $\phi = \Diamond ( \texttt{task1.1} \wedge \Diamond\, \texttt{task1.2})$, which allows the fine-tuned LLM act as a task unrelated translator, as demonstrated in~\cite{pan2023data,chen-etal-2023-nl2tl}.  Next, we ask an LLM to reinterpret these ``lifted'' LTL specifications, creating a domain-agnostic dataset containing approximately 509 unique LTL formulas and 10621 natural language descriptions produced by the LLM.


\paragraph{Action completion} 
Given an HTT, each leaf node represent a simple task on certain objects, such as ``task 1.1.1: place plates into the lower rack'' in Fig.~\ref{fig:overview}. Viewing such simple task as a sequence of action steps, LLM is asked to expand the short instruction into a sequence of pre-defined APIs. This approach helps improve alignment with robot skills and has demonstrated effectiveness~\cite{singh2023progprompt}.
For instance, the symbol $\pi_{\text{plates}}^l$ that represents task 1.1.1 can be replaced with LTL specification composed of sequential APIs: $\pi_{\text{plates}}^l = \Diamond ( \texttt{Pickup(plate)} \wedge \Diamond\, \texttt{Move(plate, lower\_rack)})$; see step 2.2 in Fig.~\ref{fig:overview}.
After this step, a complete hierarchical LTL specifications is generated.

\begin{rem}
Assuming the HTT contains $n_1$ non-leaf nodes and $n_2$ leaf nodes, our method queries the LLM $2(n_1+n_2)+1$ times. Firstly, an LLM are queried once to create the HTT without temporal relations. Subsequently, in $n_1+n_2$ times, temporal relations for non-leaf nodes and serial actions for leaf nodes are derived. Finally, nodes are tranlsated to flat LTL formulas in $n_1+n_2$ times via a fine-tuned LLM. 
\end{rem}

%% file: simulation.tex
\section{Experimental Results}
We evaluate the performance of \acronym{} both in simulated and real-world environments. For simulation, we use the AI2-THOR simulator~\cite{kolve2017ai2}, an interactive 3D environment that models various domestic settings, coupled with the ALFRED dataset~\cite{shridhar2020alfred}, which focuses on natural language comprehension and embodied actions. In real-world experiments, we arrange objects on a tabletop using single or multiple robotic arms via handover. We employ GPT-4~\cite{achiam2023gpt} and aim to answer three key questions:  
{\bf \begin{enumerate}[label=Q\arabic*.]
    \item Is \acronym{} capable of reasoning over complex human instructions effectively? 
    \item Does \acronym{} achieve higher success rates while maintaining high solution quality?   
    \item Is \acronym{} flexible enough to adjust to the verbal styles of various users?
\end{enumerate}}


\subsection{Mobile manipulation tasks in AI2-THOR}\label{sec:alfred}
\paragraph{Tasks} The ALFRED dataset contains task instructions with strictly sequential steps, which we classify as \textit{base} tasks. To create more complex tasks, we procedurally combine base tasks in same scenes to generate \textit{derivative} tasks. Specifically, the tasks are firstly identified by a LLM to ensure the same object is not included in multiple base tasks simultaneously. The base tasks that involve distinct objects are then randomly combined with randomly generated temporal relationships. Subsequently, the randomly combined tasks are then reformulated into \textit{derivative} tasks by the LLM to align more naturally with human expression patterns. The number of base tasks, varied from 1 to 4, are used to reflect the complexity of the derivative task. 50 derivative tasks are generated for each category; and one of them is shown in Fig.~\ref{fig:ai2thor}.  
We then assign 1, 2, or 4 robots, each with randomly chosen initial positions within the floor plan, leading to $4 \times 50 \times 3 = 600$ test scenarios. For simultaneous task allocation and planning, a search-based planner~\cite{luo2024simultaneous} for a multi-robot system is employed.

\begin{figure}[!t]
    \centering
    \includegraphics[width=\linewidth]{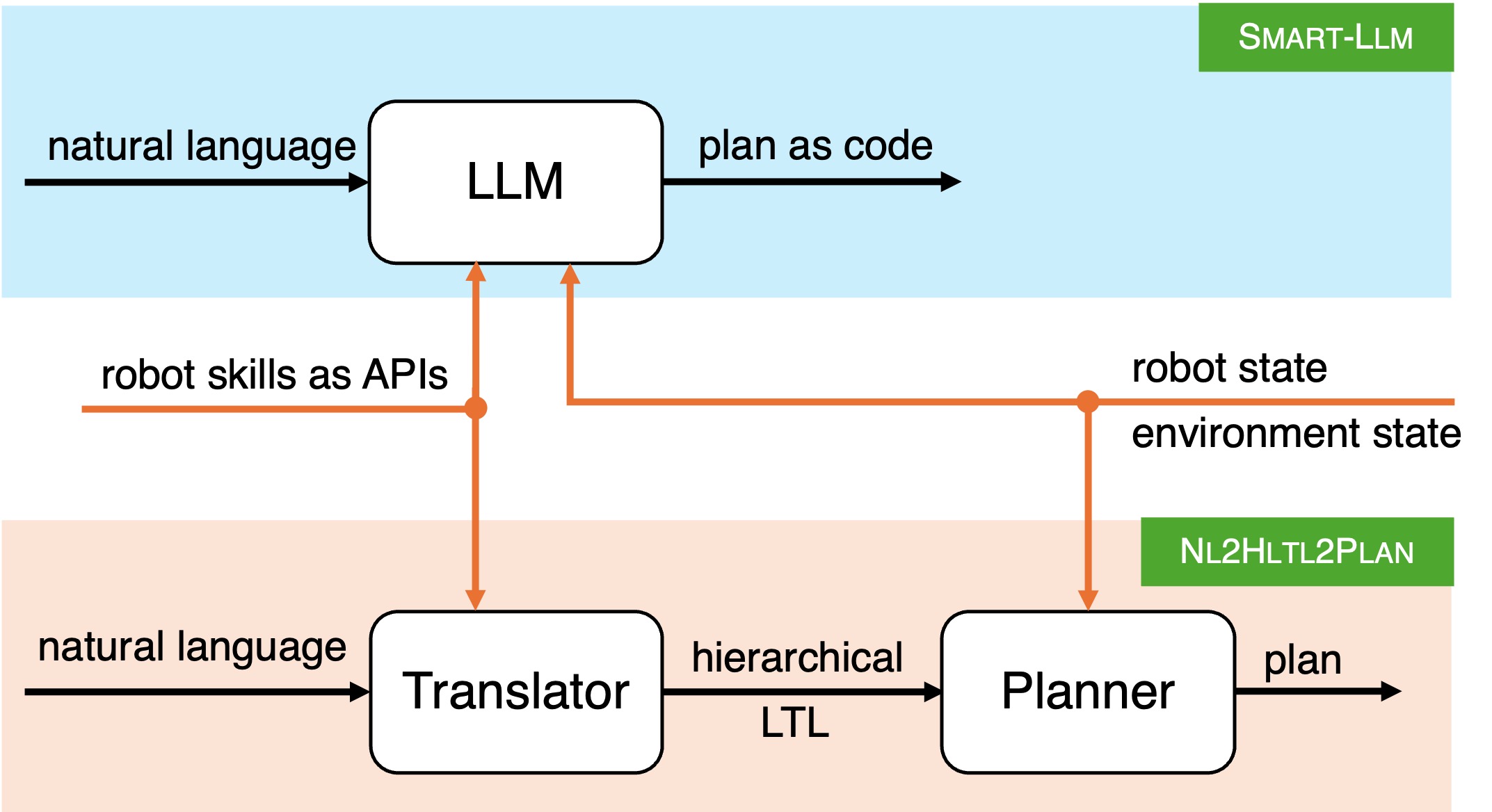}
    \caption{Comparison of pipelines from natural language to plans between \acronymsmartllm{} and \acronym.}
    \label{fig:different_pipelines}
    \vspace{-10pt}
\end{figure}

\paragraph{Comparison} We compare our method with \acronymsmartllm~\cite{kannan2023smart}, which uses an LLM to generate Python scripts invoking predefined APIs of actions for the purposes of task decomposition and task allocation. The diagram comparison of these two pipelines is displayed in Fig.~\ref{fig:different_pipelines}. Other approaches, such as those based on PDDL or LTL, face significant challenges in solving the tasks discussed in this paper. Translating instructions into PDDL fails to account for temporal constraints, while methods that expand \textit{derivative} tasks into flat LTL representations become excessively complex and are therefore unsuitable for managing the tasks presented here. 

\paragraph{Metrics} We consider the following metrics. 1) Success rate, which measures whether the target goal states of objects are achieved and if the order in which these states occur satisfies the specified temporal requirements. For a detailed analysis, we further break it down into two separate components: a) conversion, b) planning. 
2) Travel cost, measured in meters, is defined as the total distance traveled by all robots, assuming no movements in manipulation. 3) Completion time, quantified as the number of discrete time steps used to complete the tasks.

\renewcommand{\arraystretch}{1.2} 
\begin{table}[!t]
    \centering\scriptsize
    \setlength{\tabcolsep}{3pt} 
    \begin{tabular}{ccccccc}
    \bhline
     Tasks     &  \multicolumn{2}{c}{Success rate $\%$} &  \multicolumn{2}{c}{Travel cost (m)}  &  \multicolumn{2}{c}{Completion time} \\
      -robots   & ours (HLTL, PL) &\cite{kannan2023smart} & ours & \cite{kannan2023smart} & ours  & \cite{kannan2023smart}\\
    \bhline
     1-1 & {\bf 92} (100,92) &  48  & {\bf 2.9$\pm$2.0} &  4.7$\pm$3.5 & {\bf 11.5$\pm$7.9}  & 14.1$\pm$9.6 \\
      1-2  & {\bf 90} (100, 90)  & 36 & {\bf 2.7$\pm$1.9}   & 6.3$\pm$4.7 &  {\bf 10.6$\pm$8.0}  & 17.8$\pm$14.1\\
      1-4  & {\bf 88} (100, 88)   & 22 & {\bf 2.4$\pm$2.0}  & 6.0$\pm$5.9 & {\bf 9.3$\pm$7.8}   &  18.5$\pm$22.4\\
     \hline
     2-1 & {\bf 84} (92, 91)  & 16 &  {\bf 7.1$\pm$6.2}  & 10.1$\pm$4.8 & {\bf 28.4$\pm$24.8} & 28.8$\pm$16.9\\
     2-2  & {\bf 82} (92, 89)  & 10 & {\bf 7.0$\pm$6.2} & 7.3$\pm$5.0 & {\bf 22.7$\pm$21.2} & 29.0$\pm$19.8\\
     2-4  & {\bf 74} (92, 81)  & 8 & { 5.9$\pm$5.7 }&{\bf 3.8$\pm$3.3} & { 18.6$\pm$17.7} & {\bf 15.2$\pm$13.1}\\
    \hline     
    3-1 & {\bf 78} (86, 91)  & 46 & { 13.1$\pm$8.6} & {\bf 12.7$\pm$5.9} & { 52.5$\pm$34.4} & {\bf41.0$\pm$22.3}\\
     3-2  & {\bf 76} (86, 88) & 38 & {\bf 12.6$\pm$8.3} & 16.8$\pm$10.3 & {\bf 35.0$\pm$23.0}  & 50.4$\pm$43.1\\
     3-4  & {\bf 68} (86, 79)& 18 & {\bf 12.4$\pm$9.0} & 14.1$\pm$7.2 & {\bf 35.0$\pm$25.8}  &  41.6$\pm$17.7\\
    \hline
    4-1 & {\bf 74} (84, 88) & 36 & {\bf 12.5$\pm$8.3}  & 21.8$\pm$5.1 &  {\bf 50.1$\pm$33.1} & 70.9$\pm$35.6\\
     4-2  & {\bf 74} (84, 88)  & 26 & {{ 14.0$\pm$9.4}}  & {\bf 13.0$\pm$4.8} & {\bf 38.1$\pm$26.8}  & {71.5$\pm$33.5}\\
     4-4  & {\bf 64} (84, 77)  & 18 & {{\bf 12.6$\pm$7.8}}  & 13.2$\pm$1.3 & {{\bf 33.7$\pm$24.8}} &  81.8$\pm$31.6\\
    \bhline
    \end{tabular}
    \caption{Performance comparison. The success rate column first presents the overall success rate, with the success rates for conversion and planning in parentheses.}
    \label{tab:ai2thor}
    \vspace{-10pt}
\end{table}

\paragraph{Results} The dimensions of grid maps range from (25$\sim$30)$\times$(25$\sim$30) based on scene size. The statistical results are shown in Tab.~\ref{tab:ai2thor}, which are organized based on the number of base tasks included in the derivative tasks. This provides affirmative answers to our first two questions {\bf Q1} and {\bf Q2}.
\acronymsmartllm{} is limited to solving derivative tasks with only one base task, whereas our method can handle up to 4 tasks. For tasks comprising more than two base tasks, \acronymsmartllm's output exceeds the context window of GPT-4 (as its reasoning relies on the whole context), indicating that it uses a significant number of tokens to generate Python scripts. To address this, we introduced an additional layer atop \acronymsmartllm, providing a satisfying sequence of base tasks decomposed from derivative tasks. Each base task is then sequentially processed through \acronymsmartllm{} to obtain a viable solution.
In general, our approach not only achieves a higher success rate but also results in plans that are more cost-effective and require shorter amount of time to complete. For derivative tasks comprising 4 base tasks,~\acronymsmartllm{} exhibits a considerably lower success rate. However,~\acronym{} still attains a success rate of approximately 84$\%$ when converting to hierarchical LTL. As the number of robots increases, both travel costs and completion times decrease due to the parallel execution of base tasks. However, the success rate slightly decreases during the planning phase when more robots are involved as the off-the-shelf planning search time exceeds the five-minute timeout. 
We hypothesize that the planner~\cite{luo2024simultaneous} employs a best-first search strategy to ensure optimality, facing a substantial challenge due to the vast search space involving long-horizon tasks, action spaces (both navigation and manipulation), and map dimensions. More robots can be handled by upgrading to a more capable downstream planner. This flexibility in utilizing off-the-shelf planners differentiates our approach from existing studies where LLMs are primarily used for task allocation. Note that only travel cost and completion time for successfully completed tasks are recorded. Therefore, the data for \acronymsmartllm{} are not fully representative due to its lower success rate. Tasks of higher complexity, which typically involve greater travel costs and longer completion times, are more likely to fail and are thus excluded from the data. A series of snapshots capturing task execution is displayed in Fig.~\ref{fig:ai2thor}.


\subsection{Real-world rearrangement involving human participants}
The real-world tabletop experiment is with a robotic arm placing fruits and vegetables onto colored plates. Given the 2D nature of the task, we convert the environment into a discrete grid world, and use the planner~\cite{luo2024simultaneous}. The use of one arm simplifies the task compared to the multi-robot scenarios, as it eliminates task allocation.  Our evaluation has two aspects: a) the adaptability to verbal tones and styles from various users; and b) the comparative effectiveness of the plan generated from our method against existing methods. To explore the first aspect, we conduct a user study with 4 participants, asking each to rephrase the task instructions according to personal style, while maintaining the original semantics. For the second aspect, we employ an LLM as the task planner, explicitly prompting it to minimize trajectory length based on the provided initial 2D coordinates of all objects and robotic arms. This approach directly generates a sequence of API calls, similar to the method used in \textsc{ProgPrompt}~\cite{singh2023progprompt}.
We developed a dataset containing  instructions for eight arrangement tasks, each specified with temporal constraints. To address the probabilistic behavior of the LLM, we conducted 5 queries to the LLM for each rephrased instruction, resulting in a total of 25 test cases per task. In each scenario, object locations are randomized. The cost metric used is the projected travel distance of the robotic arm within a 2D space. 
\begin{figure}[!t]
    \centering
    \includegraphics[width=\linewidth]{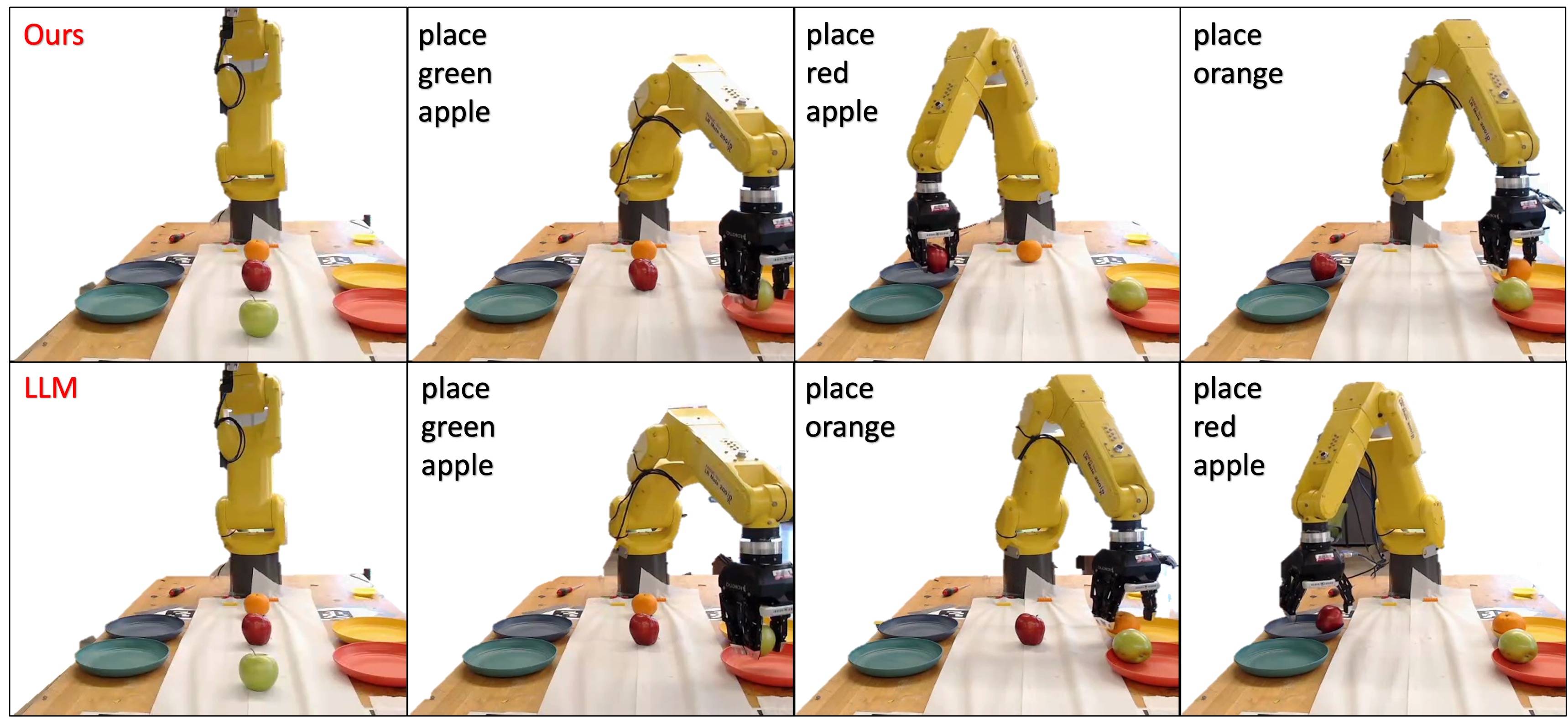}
    \caption{Comparative snapshots between \acronym{} and an LLM for task 6.~\acronym{} generates an optimal trajectory, whereas the LLM follows the sequence in which the fruits are mentioned in the instructions.}
    \label{fig:real-world-setup}
    \vspace{-10pt}
\end{figure}
\begin{table}[!t]
    \centering\scriptsize
    \begin{tabular}{lcccccc}
        \bhline
         Task &  \multicolumn{2}{c}{Success rate (\%)}  &  \multicolumn{2}{c}{Travel cost} & \multicolumn{2}{c}{Runtimes (s)} \\
         \cline{2-7} 
         ID & ours & LLM & ours & LLM & ours & LLM\\ 
         \bhline
       {\scriptsize 1} & 100  & 100  &111.2$\pm$21.6 &111.2$\pm$21.6 & 5.9$\pm$0.5   & {\bf 3.5$\pm$0.7} \\ 
       {\scriptsize 2} &100  &100  &{\bf 150.6$\pm$26.2}  &160.7$\pm$20.4 & 7.1$\pm$0.5  &  {\bf 6.7$\pm$3.2}\\ 
       {\scriptsize 3} &100  &100  &{\bf 172.5$\pm$36.4}  &211.3$\pm$27.8  & 11.5$\pm$0.3   & {\bf 5.4$\pm$0.3}\\ 
       {\scriptsize 4} & 100  & 100  &{\bf 232.7$\pm$37.8}  &235.3$\pm$35.4  &  25.7$\pm$2.4 & {\bf 6.4$\pm$0.4}\\ 
       \bhline
    \end{tabular}
    
    \vspace{3pt}  

    \begin{tabular}{p{0.2cm}p{7.6cm}}
        \bhline
     ID & \multicolumn{1}{c}{Task description} \\
     \midrule
      1   &   Place the green apple in the blue plate and the orange in the yellow plate. \\
      2   &   Place the green apple in the pink plate, the orange in the yellow plate and the red apple in the blue plate. The order of placement is not specified and can be chosen freely. \\
      3   &   Place the carrot in the blue plate, the orange in the yellow plate, the green apple in the green plate, and the red apple in the pink plate in any order. \\
      4   &   Begin by placing the green apple and orange in the yellow plate. Once done, place the carrot and red apple in the blue plate also in any order. \\
     \bhline
    \end{tabular}
    \caption{Statistical results from tabletop experiments.}
    \label{tab:tableset}
    \vspace{-10pt}
\end{table}

 \begin{figure}[!t]
    \centering
     \subfigure[Straight line configuration]{
      \label{fig:line}
      \includegraphics[width=0.38\linewidth]{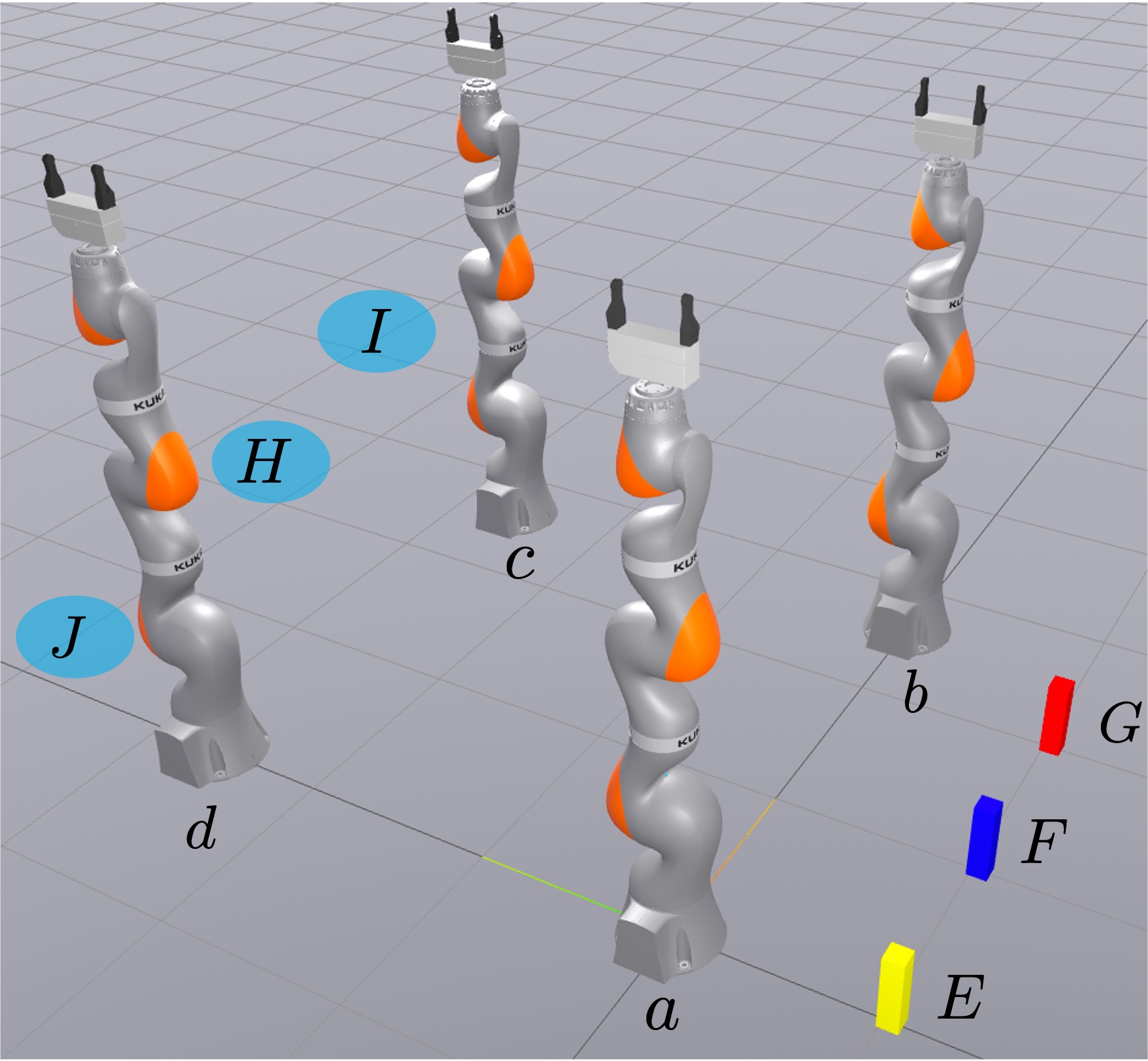}}
     \subfigure[Square configuration]{
      \label{fig:square}
      \includegraphics[width=0.58\linewidth]{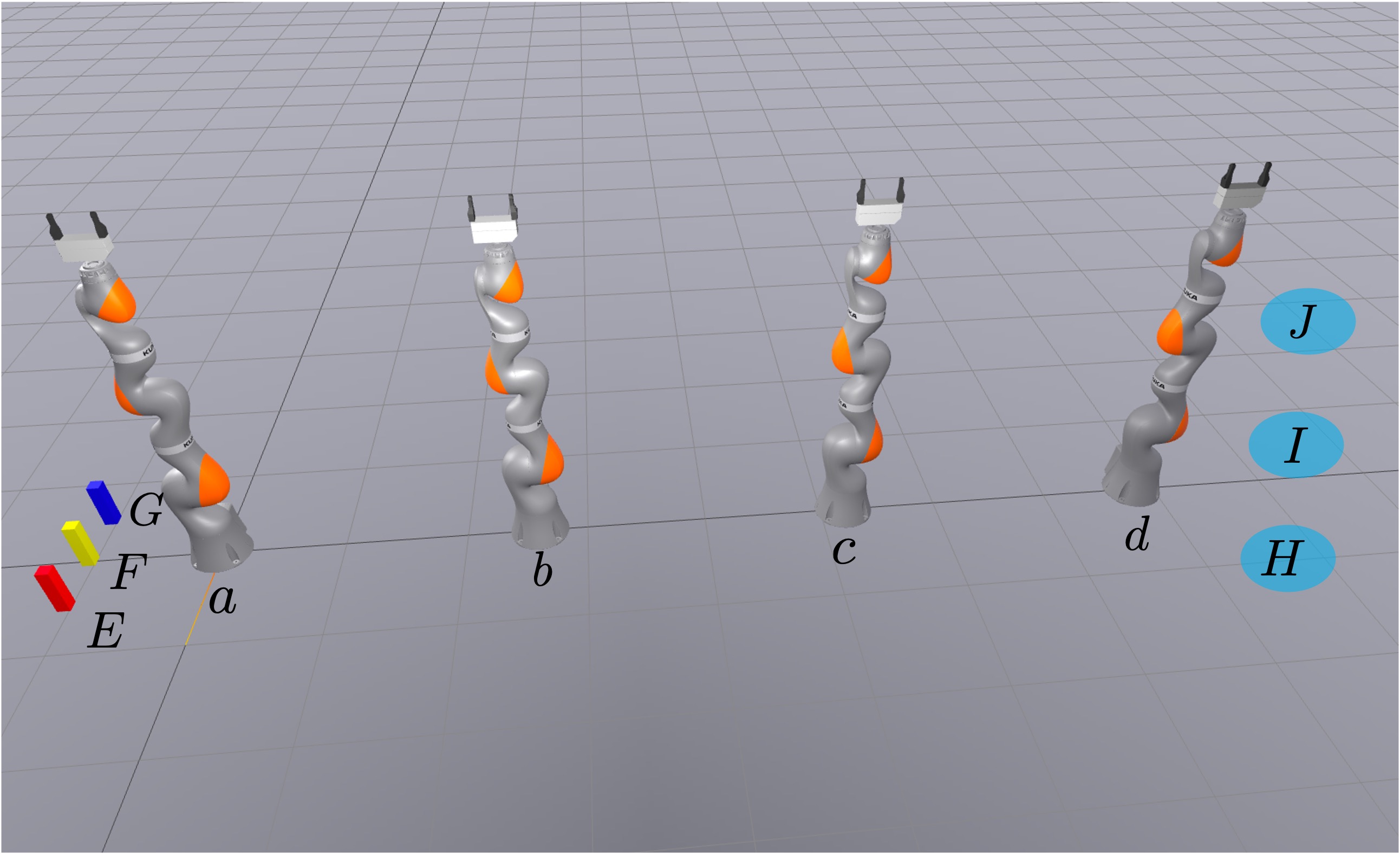}}
    \caption{Four robot arms in straight line or square configurations, where symbols $E, F$ and $G$ represent source locations and $H, I$ and $J$ denote target locations.}
    \label{fig:four_arm_config}
 \end{figure}

\begin{table}[!t]
    \centering\scriptsize
   \begin{tabular}{p{2.6cm}cccccccc}
        \bhline
         \multicolumn{1}{c}{ \multirow{2}{*}{ Success rate (\%)} } & \multicolumn{8}{c}{Task ID} \\
        \cline{2-9}
        & 1 & 2 & 3 & 4 & 5 & 6 & 7 & 8 \\
        \midrule
       \multicolumn{1}{c}{Ours}  & {\bf  100} & {\bf 100} & {\bf 80} & {\bf 80} & {\bf 80} & {\bf 90} & {\bf 100} & {\bf 100} \\
          \multicolumn{1}{c}{LLM} & 90 & 30 & 20 & 20 & 0 & 40 & 10 & 40 \\
        \bhline
    \end{tabular}

    \vspace{4pt}

    \begin{tabular}{p{0.2cm}p{7.8cm}}
        \bhline
         ID &	\multicolumn{1}{c}{task description} \\
        \midrule
        1 &	Transfer three blocks into the opposite. The red block should be placed before the blue block. \\
        2 &	Put the three blocks into three goal region at region H at, region I and region J each goal region can only be placed with one object. \\
        3 &	Please plan a solution to move the three pillars to the placement area (region E, F, G) to the side of the arm a and arm b. Each object cannot be placed on the area other than the target region. \\
        4 &	Transfer three blocks into the opposite regions. The robot arm cannot touch blue block until any of the robot arm touched the red block. After all of these, you should place yellow block back to region G (where the red block is currently locating). \\
        5 &	The red block should be placed before the blue block. Please exchange the position of the red block and blue block. At any time, put the yellow block into the opposite side. \\
        6 &	The red block should be placed before the blue block. Please place the three objects separately to region G, H and I. \\
        7 &	The red block should be placed before the blue block. Please place the three objects one after another to region I, no objects can be placed on region F, G and H during this period. \\
        8 &	Please first place the yellow block to region I, then place the yellow block to region G, and move red and blue blocks to any empty region at any time.  \\
        \bhline
    \end{tabular}
    \caption{Statistical results for multi-robot handover. The first five tasks involve scenarios where four arms are arranged in a square formation, while the last three tasks involve scenarios where the four arms are aligned in a straight line.}
    \label{tab:multi_arm_handover}
\end{table}

The results are presented in Tab.~\ref{tab:tableset}, which positively answers {\bf Q3}. As observed, both~\acronym{} and the LLM achieve a high success rate, which aligns with the expectations given the task complexities. 
Regarding cost, with multiple feasible solutions,~\acronym{} consistently produces lower-cost paths, with the exception of task 1. In this task, the LLM manages to create an optimal plan given the placement of fruits. 
The runtimes include the time to obtain the executable action sequence.~\acronym{} experienced slightly longer runtimes compared to the LLM because querying times varies in different HTT structure. Comparison between~\acronym{} and the LLM is displayed in Fig.~\ref{fig:real-world-setup}. 

\subsection{Multi-robot handover tasks}\label{sec:multi-robot}
We examine the execution of pick-and-place tasks involving multiple objects by four fixed robot arms, which are either aligned in a straight line or arranged in a square configuration; see Fig.~\ref{fig:four_arm_config}. Certain tasks might necessitate the transfer of objects between robots, depending on their proximity. The planner our approach uses, inspired from work~\cite{kurtz2023temporal}, produces collision-free trajectories by simultaneously considering task and motion planning. The prompt for the baseline that directly uses the LLM as task planner is illustrated in Fig.~\ref{fig:overview}. 

\begin{figure}[!t]
    \centering
    \includegraphics[width=\linewidth]{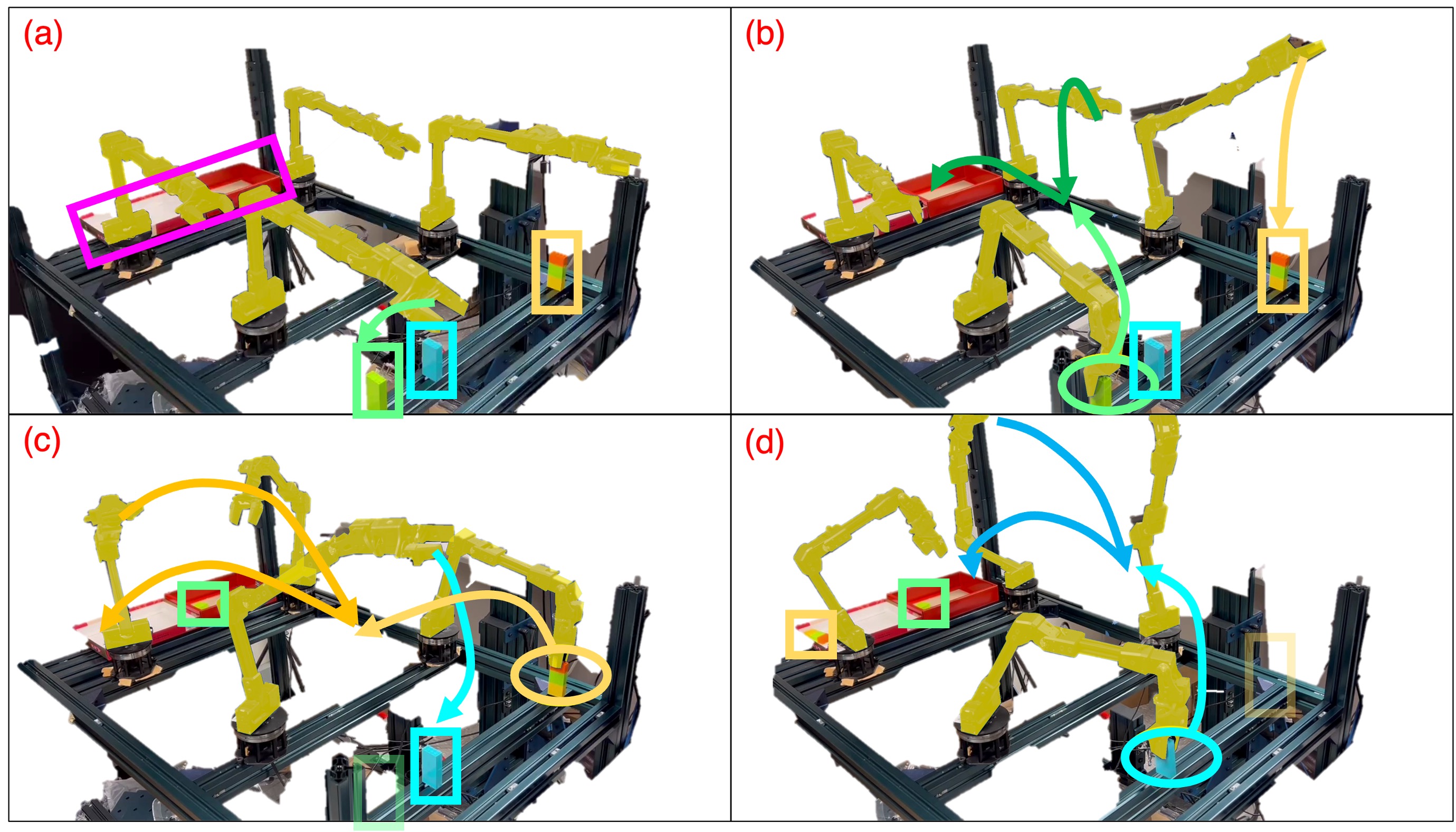}
    \caption{Snapshots depict four arms performing real-world tasks of picking and placing objects via handover. The instruction given is, ``Please move the blue, green, and multi-colored blocks to the two opposite boxes, place the colored ones after the green ones.'' Target areas are colored in magenta. The block being selected is emphasized with an ellipse, and the remaining blocks are contained within rectangles. The colored curves with arrows illustrate the trajectories of the end-effectors, where head-to-head arrows indicate handovers between robotic arms.
    }
    \label{fig:real-world-setup_four_arm}
    \vspace{-10pt}
\end{figure}

Tab.~\ref{tab:multi_arm_handover} displays eight multi-stage pick-and-place tasks with temporal constraints. For the LLM-based planner, a planning scheme is deemed successful if it allows for the sequential actions of multiple robots to be executed successfully while adhering to the temporal constraints. It is evident that for tasks involving robot handovers, the success rate of the LLM-based planner decreases due to the need for cooperative planning. Considering the probabilistic output of GPT-4, we conducted 10 tests per task to enhance the diversity of the LLM's responses. The results indicate that by dividing the planning process into task hierarchical extraction and LTL-based optimization, we can effectively bypass direct control of robots' low-level movements, thereby improving completion of multi-stage handover tasks. Moreover, we conduct experiments in a real-world setting with four robotic arms, and a series of snapshots are presented in  Fig.~\ref{fig:real-world-setup_four_arm}.

\subsection{Analysis of failure reasons}

We categorize the causes of failure into four groups, each aligned with the four stages described in Sections~\ref{sec:htt} and~\ref{sec:translation}. These failure rates are presented in Tab.~\ref{tab:failure_cases}.

\paragraph{Task decomposition} The breakdown of tasks might omit certain subtasks. For instance, a decomposition of ``Heat a sliced tomato in microwave.'' could be ``1.1 slice a tomato, 1.2 heat the tomato in microwave''. However, there is a necessary intermediate step absent between 1.1 and 1.2, which should be \underline{1.2 place the tomato in microwave}.


\paragraph{Temporal extraction} Ambiguous wording might cause an LLM to recognize only a subset of temporal relations. Consider the sequence: ``Put a bread in the oven [1.1], place a pot in the pool [1.2]. At any time, move a bowl to the desk [1.3].'' The output suggests 1.2 and 1.3 can occur in any order, and \underline{1.2 follows 1.1}. This inference arises from the absence of clear directives on order. A better inference would indicate that 1.1 and 1.2 can be performed in any order.

\paragraph{LTL translation} The conversion process may result in inaccuracies. For example, the specified task sequence “First 1.1, then 1.2, 1.3, and 1.4 can be completed in any sequence” is erroneously translated as $\Diamond ( p_{1.1} \wedge ( \Diamond ( p_{1.2} \wedge ( p_{1.3} \wedge p_{1.4} ) ) ) )$. The correct formula should be $\Diamond ( p_{1.1} \wedge \Diamond  p_{1.2} \wedge \Diamond p_{1.3} \wedge \Diamond p_{1.4} )$.

\paragraph{Action completion} Redundant actions that duplicate previous ones may occur. For instance, the phrase “Place \textbf{the} sliced tomato on the pan,” which functions as a leaf node in HTT, implies that the tomato has already been sliced. A redundant sequence like “\underline{pick(tomato), slice(tomato)}, put(pan, tomato)” would be inappropriate here, as it reflects the actions for a non-leaf node, such as “Place \textbf{a} sliced tomato on the pan.”

\begin{table}[!t]
\centering\scriptsize
\begin{tabular}{ccccc}
\bhline
\multirow{2}{*}{Error type}  &  task & temporal  & LTL  & action  \\
  & decomposition & extraction & translation & completion\\ 
\midrule
Failure rate \% & 1.33 & 1.83 & 2.83  & 2.50 \\
\bhline
\end{tabular}
    \caption{Statistics of failure cases. The results derive from scenarios in Sections~\ref{sec:alfred} and ~\ref{sec:multi-robot}, encompassing 208 task descriptions altogether. The rate is determined across the total instances in the HTT. An HTT comprising $n$ non-leaf nodes and $m$ leaf nodes accounts for $m$ instances in action completion and $n$ instances across the other three categories.}
    \label{tab:failure_cases}
    \vspace{-10pt}
\end{table}

%% file: conclusion.tex
\section{Conclusions and Limitations}\label{sec:conclusion}
We proposed~\acronym{} to transform unstructured language into a structured, hierarchical formal representation--hierarchical LTL, where the lowest level corresponds to sequentially ordered robot actions. The task representation is ready to be used by off-the-shelf planners for multi-robot systems. Our simulation and real-world experiment outcomes demonstrated that the framework offers an intuitive and user-friendly approach for deploying robots in daily situations. 

{\bf Limitations:} The~\acronym{} operates as an open loop without feedback. To transition to a closed-loop one, it is essential to integrate a syntax checker and a semantic checker. The syntax checker verifies adherence to the hierarchical LTL structure necessary for HTT representation. Meanwhile, the semantic checker offers feedback on errors when the planner fails to identify a solution. Another limitation is that once created, the HTT representation remains unchanged. We derive an LTL specification by extracting child tasks from a parent task. As more child tasks are included, the accuracy of translation drops. Therefore, to handle tasks with more base tasks, it is necessary to restructure the HTT to restrict the number of child tasks a single parent task has.

%% file: root.bbl
\begin{thebibliography}{10}
\providecommand{\url}[1]{#1}
\csname url@rmstyle\endcsname
\providecommand{\newblock}{\relax}
\providecommand{\bibinfo}[2]{#2}
\providecommand\BIBentrySTDinterwordspacing{\spaceskip=0pt\relax}
\providecommand\BIBentryALTinterwordstretchfactor{4}
\providecommand\BIBentryALTinterwordspacing{\spaceskip=\fontdimen2\font plus
\BIBentryALTinterwordstretchfactor\fontdimen3\font minus \fontdimen4\font\relax}
\providecommand\BIBforeignlanguage[2]{{%
\expandafter\ifx\csname l@#1\endcsname\relax
\typeout{** WARNING: IEEEtran.bst: No hyphenation pattern has been}%
\typeout{** loaded for the language `#1'. Using the pattern for}%
\typeout{** the default language instead.}%
\else
\language=\csname l@#1\endcsname
\fi
#2}}

\bibitem{padalkar2023open}
A.~Padalkar, A.~Pooley, A.~Jain, A.~Bewley, A.~Herzog, A.~Irpan, A.~Khazatsky, A.~Rai, A.~Singh, A.~Brohan, \emph{et~al.}, ``Open x-embodiment: Robotic learning datasets and rt-x models,'' \emph{arXiv preprint arXiv:2310.08864}, 2023.

\bibitem{belle2023neuro}
V.~Belle, M.~Fisher, A.~Russo, E.~Komendantskaya, and A.~Nottle, ``Neuro-symbolic ai+ agent systems: A first reflection on trends, opportunities and challenges,'' in \emph{International Conference on Autonomous Agents and Multiagent Systems}.\hskip 1em plus 0.5em minus 0.4em\relax Springer, 2023, pp. 180--200.

\bibitem{cohen2024survey}
V.~Cohen, J.~X. Liu, R.~Mooney, S.~Tellex, and D.~Watkins, ``A survey of robotic language grounding: Tradeoffs between symbols and embeddings,'' \emph{arXiv preprint arXiv:2405.13245}, 2024.

\bibitem{tenenbaum2011grow}
J.~B. Tenenbaum, C.~Kemp, T.~L. Griffiths, and N.~D. Goodman, ``How to grow a mind: Statistics, structure, and abstraction,'' \emph{science}, vol. 331, no. 6022, pp. 1279--1285, 2011.

\bibitem{kemp2007learning}
C.~Kemp, A.~Perfors, and J.~B. Tenenbaum, ``Learning overhypotheses with hierarchical bayesian models,'' \emph{Developmental science}, vol.~10, no.~3, pp. 307--321, 2007.

\bibitem{liembodied}
M.~Li \emph{et~al.}, ``Embodied agent interface: Benchmarking llms for embodied decision making,'' in \emph{The Thirty-eight Conference on Neural Information Processing Systems Datasets and Benchmarks Track}.

\bibitem{luo2024simultaneous}
X.~Luo and C.~Liu, ``Simultaneous task allocation and planning for multi-robots under hierarchical temporal logic specifications,'' \emph{arXiv preprint arXiv:2401.04003}, 2024.

\bibitem{luo2021abstraction}
X.~Luo, Y.~Kantaros, and M.~M. Zavlanos, ``An abstraction-free method for multirobot temporal logic optimal control synthesis,'' \emph{IEEE Transactions on Robotics}, vol.~37, no.~5, pp. 1487--1507, 2021.

\bibitem{luo2022temporal}
X.~Luo and M.~M. Zavlanos, ``Temporal logic task allocation in heterogeneous multirobot systems,'' \emph{IEEE Transactions on Robotics}, vol.~38, no.~6, pp. 3602--3621, 2022.

\bibitem{xu2023large}
F.~Xu, Q.~Lin, J.~Han, T.~Zhao, J.~Liu, and E.~Cambria, ``Are large language models really good logical reasoners? a comprehensive evaluation from deductive, inductive and abductive views,'' \emph{arXiv preprint arXiv:2306.09841}, 2023.

\bibitem{liu2023grounding}
J.~X. Liu, Z.~Yang, I.~Idrees, S.~Liang, B.~Schornstein, S.~Tellex, and A.~Shah, ``Grounding complex natural language commands for temporal tasks in unseen environments,'' in \emph{7th Annual Conference on Robot Learning}, 2023.

\bibitem{ghallab2016automated}
M.~Ghallab, D.~Nau, and P.~Traverso, \emph{Automated planning and acting}.\hskip 1em plus 0.5em minus 0.4em\relax Cambridge University Press, 2016.

\bibitem{chen2023nl2tl}
Y.~Chen, R.~Gandhi, Y.~Zhang, and C.~Fan, ``Nl2tl: Transforming natural languages to temporal logics using large language models,'' \emph{arXiv preprint arXiv:2305.07766}, 2023.

\bibitem{cosler2023nl2spec}
M.~Cosler, C.~Hahn, D.~Mendoza, F.~Schmitt, and C.~Trippel, ``nl2spec: Interactively translating unstructured natural language to temporal logics with large language models,'' in \emph{International Conference on Computer Aided Verification}.\hskip 1em plus 0.5em minus 0.4em\relax Springer, 2023, pp. 383--396.

\bibitem{team2024octo}
O.~M. Team, D.~Ghosh, H.~Walke, K.~Pertsch, K.~Black, O.~Mees, S.~Dasari, J.~Hejna, T.~Kreiman, C.~Xu, \emph{et~al.}, ``Octo: An open-source generalist robot policy,'' \emph{arXiv preprint arXiv:2405.12213}, 2024.

\bibitem{jiang2022vima}
Y.~Jiang, A.~Gupta, Z.~Zhang, G.~Wang, Y.~Dou, Y.~Chen, L.~Fei-Fei, A.~Anandkumar, Y.~Zhu, and L.~Fan, ``Vima: General robot manipulation with multimodal prompts,'' in \emph{NeurIPS 2022 Foundation Models for Decision Making Workshop}, 2022.

\bibitem{li2023vision}
X.~Li \emph{et~al.}, ``Vision-language foundation models as effective robot imitators,'' \emph{arXiv preprint arXiv:2311.01378}, 2023.

\bibitem{xie2023translating}
Y.~Xie, C.~Yu, T.~Zhu, J.~Bai, Z.~Gong, and H.~Soh, ``Translating natural language to planning goals with large-language models,'' \emph{arXiv preprint arXiv:2302.05128}, 2023.

\bibitem{liu2023llm+}
B.~Liu, Y.~Jiang, X.~Zhang, Q.~Liu, S.~Zhang, J.~Biswas, and P.~Stone, ``Llm+ p: Empowering large language models with optimal planning proficiency,'' \emph{arXiv preprint arXiv:2304.11477}, 2023.

\bibitem{valmeekam2024planbench}
K.~Valmeekam, M.~Marquez, A.~Olmo, S.~Sreedharan, and S.~Kambhampati, ``Planbench: An extensible benchmark for evaluating large language models on planning and reasoning about change,'' \emph{Advances in Neural Information Processing Systems}, vol.~36, 2024.

\bibitem{singh2023progprompt}
I.~Singh, V.~Blukis, A.~Mousavian, A.~Goyal, D.~Xu, J.~Tremblay, D.~Fox, J.~Thomason, and A.~Garg, ``Progprompt: program generation for situated robot task planning using large language models,'' \emph{Autonomous Robots}, pp. 1--14, 2023.

\bibitem{liang2023code}
J.~Liang, W.~Huang, F.~Xia, P.~Xu, K.~Hausman, B.~Ichter, P.~Florence, and A.~Zeng, ``Code as policies: Language model programs for embodied control,'' in \emph{2023 IEEE International Conference on Robotics and Automation (ICRA)}.\hskip 1em plus 0.5em minus 0.4em\relax IEEE, 2023, pp. 9493--9500.

\bibitem{huang2023voxposer}
W.~Huang, C.~Wang, R.~Zhang, Y.~Li, J.~Wu, and L.~Fei-Fei, ``Voxposer: Composable 3d value maps for robotic manipulation with language models,'' in \emph{Conference on Robot Learning}.\hskip 1em plus 0.5em minus 0.4em\relax PMLR, 2023, pp. 540--562.

\bibitem{kannan2023smart}
S.~S. Kannan, V.~L. Venkatesh, and B.-C. Min, ``Smart-llm: Smart multi-agent robot task planning using large language models,'' \emph{arXiv preprint arXiv:2309.10062}, 2023.

\bibitem{hu2024deploying}
Z.~Hu, F.~Lucchetti, C.~Schlesinger, Y.~Saxena, A.~Freeman, S.~Modak, A.~Guha, and J.~Biswas, ``Deploying and evaluating llms to program service mobile robots,'' \emph{IEEE Robotics and Automation Letters}, 2024.

\bibitem{wang2023voyager}
G.~Wang, Y.~Xie, Y.~Jiang, A.~Mandlekar, C.~Xiao, Y.~Zhu, L.~Fan, and A.~Anandkumar, ``Voyager: An open-ended embodied agent with large language models,'' \emph{arXiv preprint arXiv:2305.16291}, 2023.

\bibitem{brohan2023can}
A.~Brohan, Y.~Chebotar, C.~Finn, K.~Hausman, A.~Herzog, D.~Ho, J.~Ibarz, A.~Irpan, E.~Jang, R.~Julian, \emph{et~al.}, ``Do as i can, not as i say: Grounding language in robotic affordances,'' in \emph{Conference on robot learning}.\hskip 1em plus 0.5em minus 0.4em\relax PMLR, 2023, pp. 287--318.

\bibitem{huang2023inner}
W.~Huang \emph{et~al.}, ``Inner monologue: Embodied reasoning through planning with language models,'' in \emph{Conference on Robot Learning}.\hskip 1em plus 0.5em minus 0.4em\relax PMLR, 2023, pp. 1769--1782.

\bibitem{ren2023robots}
A.~Z. Ren, A.~Dixit, A.~Bodrova, S.~Singh, S.~Tu, N.~Brown, P.~Xu, L.~Takayama, F.~Xia, J.~Varley, \emph{et~al.}, ``Robots that ask for help: Uncertainty alignment for large language model planners,'' in \emph{7th Annual Conference on Robot Learning}, 2023.

\bibitem{konrad2005real}
S.~Konrad and B.~H. Cheng, ``Real-time specification patterns,'' in \emph{Proceedings of the 27th international conference on Software engineering}, 2005, pp. 372--381.

\bibitem{fuggitti2023nl2ltl}
F.~Fuggitti and T.~Chakraborti, ``Nl2ltl--a python package for converting natural language (nl) instructions to linear temporal logic (ltl) formulas,'' in \emph{AAAI Conference on Artificial Intelligence}, 2023.

\bibitem{pan2023data}
J.~Pan, G.~Chou, and D.~Berenson, ``Data-efficient learning of natural language to linear temporal logic translators for robot task specification,'' in \emph{2023 IEEE International Conference on Robotics and Automation (ICRA)}.\hskip 1em plus 0.5em minus 0.4em\relax IEEE, 2023, pp. 11\,554--11\,561.

\bibitem{patel2019learning}
R.~Patel, R.~Pavlick, and S.~Tellex, ``Learning to ground language to temporal logical form,'' in \emph{Conference of the North American Chapter of the Association for Computational Linguistics (NAACL)}, 2019.

\bibitem{wang2021learning}
C.~Wang, C.~Ross, Y.-L. Kuo, B.~Katz, and A.~Barbu, ``Learning a natural-language to ltl executable semantic parser for grounded robotics,'' in \emph{Conference on Robot Learning}.\hskip 1em plus 0.5em minus 0.4em\relax PMLR, 2021, pp. 1706--1718.

\bibitem{liu2023lang2ltl}
J.~X. Liu, Z.~Yang, I.~Idrees, S.~Liang, B.~Schornstein, S.~Tellex, and A.~Shah, ``Lang2ltl: Translating natural language commands to temporal robot task specification,'' \emph{arXiv preprint arXiv:2302.11649}, 2023.

\bibitem{chen-etal-2023-nl2tl}
Y.~Chen, R.~Gandhi, Y.~Zhang, and C.~Fan, ``{NL}2{TL}: Transforming natural languages to temporal logics using large language models,'' in \emph{Proceedings of the 2023 Conference on Empirical Methods in Natural Language Processing}.\hskip 1em plus 0.5em minus 0.4em\relax Singapore: Association for Computational Linguistics, Dec. 2023, pp. 15\,880--15\,903.

\bibitem{hsu2024s}
J.~Hsu, J.~Mao, J.~Tenenbaum, and J.~Wu, ``What’s left? concept grounding with logic-enhanced foundation models,'' \emph{Advances in Neural Information Processing Systems}, vol.~36, 2024.

\bibitem{wang2023conformal}
J.~Wang \emph{et~al.}, ``Conformal temporal logic planning using large language models: Knowing when to do what and when to ask for help,'' \emph{arXiv preprint arXiv:2309.10092}, 2023.

\bibitem{mandi2023roco}
Z.~Mandi, S.~Jain, and S.~Song, ``Roco: Dialectic multi-robot collaboration with large language models,'' \emph{arXiv preprint arXiv:2307.04738}, 2023.

\bibitem{lykov2023llm}
A.~Lykov \emph{et~al.}, ``Llm-mars: Large language model for behavior tree generation and nlp-enhanced dialogue in multi-agent robot systems,'' \emph{arXiv preprint arXiv:2312.09348}, 2023.

\bibitem{chen2023scalable}
Y.~Chen, J.~Arkin, Y.~Zhang, N.~Roy, and C.~Fan, ``Scalable multi-robot collaboration with large language models: Centralized or decentralized systems?'' \emph{arXiv preprint arXiv:2309.15943}, 2023.

\bibitem{garg2024large}
K.~Garg, J.~Arkin, S.~Zhang, N.~Roy, and C.~Fan, ``Large language models to the rescue: Deadlock resolution in multi-robot systems,'' \emph{arXiv preprint arXiv:2404.06413}, 2024.

\bibitem{wang2024safe}
J.~Wang, G.~He, and Y.~Kantaros, ``Safe task planning for language-instructed multi-robot systems using conformal prediction,'' \emph{arXiv preprint arXiv:2402.15368}, 2024.

\bibitem{yu2023co}
B.~Yu, H.~Kasaei, and M.~Cao, ``Co-navgpt: Multi-robot cooperative visual semantic navigation using large language models,'' \emph{arXiv preprint arXiv:2310.07937}, 2023.

\bibitem{baier2008principles}
C.~Baier and J.-P. Katoen, \emph{Principles of model checking}.\hskip 1em plus 0.5em minus 0.4em\relax MIT press Cambridge, 2008.

\bibitem{kupferman2001model}
O.~Kupferman and M.~Y. Vardi, ``Model checking of safety properties,'' \emph{Formal methods in system design}, vol.~19, pp. 291--314, 2001.

\bibitem{petroni2019language}
F.~Petroni \emph{et~al.}, ``Language models as knowledge bases?'' in \emph{Proceedings of the 2019 Conference on Empirical Methods in Natural Language Processing and the 9th International Joint Conference on Natural Language Processing (EMNLP-IJCNLP)}, 2019, pp. 2463--2473.

\bibitem{jiang2023mistral}
A.~Q. Jiang, A.~Sablayrolles, A.~Mensch, C.~Bamford, D.~S. Chaplot, D.~d.~l. Casas, F.~Bressand, G.~Lengyel, G.~Lample, L.~Saulnier, \emph{et~al.}, ``Mistral 7b,'' \emph{arXiv preprint arXiv:2310.06825}, 2023.

\bibitem{kolve2017ai2}
E.~Kolve, R.~Mottaghi, W.~Han, E.~VanderBilt, L.~Weihs, A.~Herrasti, M.~Deitke, K.~Ehsani, D.~Gordon, Y.~Zhu, \emph{et~al.}, ``Ai2-thor: An interactive 3d environment for visual ai,'' \emph{arXiv preprint arXiv:1712.05474}, 2017.

\bibitem{shridhar2020alfred}
M.~Shridhar \emph{et~al.}, ``Alfred: A benchmark for interpreting grounded instructions for everyday tasks,'' in \emph{Proceedings of the IEEE/CVF conference on computer vision and pattern recognition}, 2020, pp. 10\,740--10\,749.

\bibitem{achiam2023gpt}
J.~Achiam, S.~Adler, S.~Agarwal, L.~Ahmad, I.~Akkaya, F.~L. Aleman, D.~Almeida, J.~Altenschmidt, S.~Altman, S.~Anadkat, \emph{et~al.}, ``Gpt-4 technical report,'' \emph{arXiv preprint arXiv:2303.08774}, 2023.

\bibitem{kurtz2023temporal}
V.~Kurtz and H.~Lin, ``Temporal logic motion planning with convex optimization via graphs of convex sets,'' \emph{IEEE Transactions on Robotics}, 2023.

\end{thebibliography}
